\documentclass[10pt,twocolumn,letterpaper]{article}

\usepackage{cvpr} 
\usepackage{times}
\usepackage{epsfig}
\usepackage{graphicx}
\usepackage{amsmath}
\usepackage{amssymb}
\usepackage{booktabs}
\usepackage{multirow}
\usepackage{subcaption}
\usepackage[table]{xcolor}
\usepackage{dsfont}

\newcommand{\rev}[1]{#1}

\makeatletter
\newcommand\blfootnote[1]{%
  \begingroup
  \renewcommand\thefootnote{}\footnote{#1}%
  \addtocounter{footnote}{-1}%
  \endgroup
}
\makeatother

\begin{document}

\title{Quality-Aware Multimodal Fusion Reveals \\Implicit Identity in Valence-Arousal Features}

\author{Jisu Kim\\
University of Nebraska-Lincoln\\
Lincoln, NE 68588\\
{\tt\small jkim73@huskers.unl.edu}
\and
Benjamin S. Riggan\\
University of Nebraska-Lincoln\\
Lincoln, NE 68588\\
{\tt\small briggan2@unl.edu}
}

\maketitle
\thispagestyle{empty}
\blfootnote{\textcopyright~2026 IEEE. Personal use of this material is permitted. Permission from IEEE must be obtained for all other uses, in any current or future media, including reprinting/republishing this material for advertising or promotional purposes, creating new collective works, for resale or redistribution to servers or lists, or reuse of any copyrighted component of this work in other works. Accepted for publication at the IEEE International Joint Conference on Biometrics (IJCB), 2026. The final published version will be available via IEEE Xplore.}

\begin{abstract}
Conventional face recognition relies on static appearance cues and degrades in unconstrained settings with expression variation, occlusion, and poor lighting. We hypothesize that audiovisual expression dynamics carry identity-discriminative information complementary to static appearance, and that extracting this signal requires multimodal representations robust to the variable input quality of in-the-wild video. To learn such representations, we cast multimodal valence-arousal (VA) estimation as a pretext task and propose Quality-Aware Adaptive Fusion (QAAF), which estimates per-sample, per-modality reliability and adapts each modality's contribution through learned soft gating and a quality-dependent dropout. For the problem of VA estimation, QAAF achieves an average Concordance Correlation Coefficient (CCC) of 0.472 via late fusion ensembling on Aff-wild2, \rev{improving over a baseline ensemble under the same setting (0.415) as well as a single-backbone baseline (0.288)}. Furthermore, the proposed QAAF demonstrates greater resilience to unavailable modalities, with only a 7.5--34.4\% relative decrease in CCC when one modality is missing. We then probe whether these VA-trained features encode identity without identity-specific training. On AFEW-VA (67 actors) and YTF (1,595 subjects), VA-trained backbone features rank first among evaluated soft biometric methods, and score-level fusion with ArcFace lowers EER on both datasets (0.022$\to$0.021 on AFEW-VA, 0.106$\to$0.104 on YTF), correcting 68.2\% of ArcFace's false accepts on AFEW-VA. These findings establish multimodal VA estimation as a soft biometric modality complementary to conventional face recognition.
\end{abstract}

\section{Introduction}
\label{sec:intro}

\begin{figure}[t]
    \centering
    \includegraphics[width=\columnwidth]{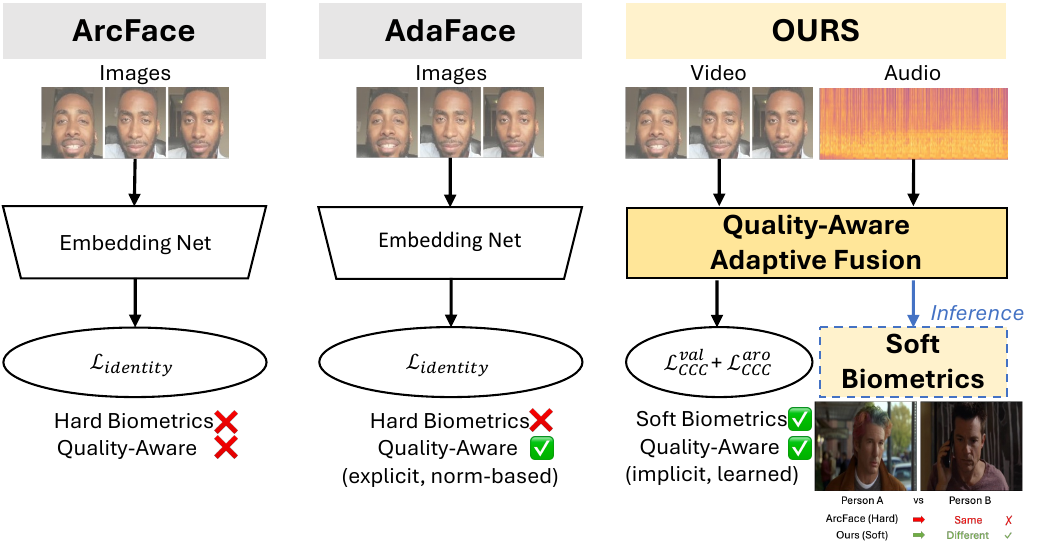}
    \caption{Overview of the proposed approach in comparison with existing face recognition paradigms. ArcFace~\cite{deng2019arcface} trains an embedding network for identity without quality awareness; AdaFace~\cite{kim2022adaface} introduces quality awareness explicitly via norm-based margins; our approach trains a multimodal model with Quality-Aware Adaptive Fusion (QAAF) on valence-arousal (VA) estimation from face video and audio, where backbone features emerge as a soft biometric signal with implicit, learned quality awareness. These features correctly distinguish impostor pairs that ArcFace false-accepts, showing that VA training implicitly encodes identity-discriminative information complementary to static appearance.}
    \label{fig:highlevel}
\end{figure}

Face recognition has saturated on standard benchmarks under controlled conditions, but its reliability degrades in unconstrained settings where expression variation, occlusion, and lighting changes alter the static appearance cues on which conventional systems depend~\cite{deng2019arcface, kim2022adaface}. This raises the question of whether other signals, such as temporal facial dynamics and vocal characteristics, can complement static appearance in such settings. Facial expression dynamics are highly individual, with temporal patterns of facial motion shown to be person-specific and persistent over time~\cite{benedikt2010facial}. The ways a person's face moves and how their voice modulates during speech reflect personal habits of articulation, emotional reactivity, and prosodic style. Building on this, prior work has demonstrated that speech-related facial motions can be used for behavioral biometrics~\cite{benedikt2010facial}, and that vocal characteristics serve as a well-established speaker recognition cue~\cite{kinnunen2010speaker}. We hypothesize that audiovisual expression dynamics, when learned jointly through a multimodal model, can serve as a soft biometric modality complementary to static appearance.

Extracting a representation that captures these audiovisual expression dynamics requires a model robust to the variable input quality of in-the-wild video. We argue that multimodal valence-arousal (VA) estimation~\cite{ringeval2013introducing, kollias2019deep}---predicting continuous emotional polarity (valence) and intensity (arousal) from visual and auditory signals---provides a suitable training signal. VA estimation forces the model to attend to expression dynamics rather than static identity cues, and multimodal fusion exploits complementary information from both modalities, consistently outperforming unimodal approaches~\cite{tzirakis2017end}.

However, a critical challenge in multimodal VA estimation is that the fidelity with which each modality captures emotion-relevant information varies significantly across samples and recording conditions, due to environmental noise, occlusion, motion blur, or poor lighting. Existing approaches such as uniform modality dropout~\cite{neverova2015moddrop}, energy-based quality estimation~\cite{zhang2023qmf}, and missing-modality reconstruction~\cite{ma2021smil, zhao2023missmodal, nezakati2024mmp} address this only partially, and none combine quality-aware gating with adaptive regularization to handle both inference-time reliability estimation and training-time robustness.

To address these limitations, we propose \textbf{Quality-Aware Adaptive Fusion (QAAF)}, a framework that estimates per-sample, per-modality reliability and dynamically adjusts each modality's contribution to the fused representation. QAAF introduces two complementary mechanisms: (1) a learned soft gating module that scales modality contributions based on estimated reliability during both training and inference and (2) a quality-dependent dropout strategy that provides adaptive regularization during training by dropping less reliable modalities more frequently. Combining these mechanisms enables the model to use quality information at inference time and learn robust single-modality representations during training. Fig.~\ref{fig:highlevel} illustrates the proposed approach in comparison with ArcFace~\cite{deng2019arcface} (no quality awareness) and AdaFace~\cite{kim2022adaface} (explicit, norm-based quality awareness). Our approach trains a multimodal model on VA estimation, where backbone features emerge as a soft biometric signal with implicit, learned quality awareness that complements face recognition.

The contributions of this work are:
\begin{enumerate}
    \item We propose Quality-Aware Adaptive Fusion (QAAF) and use it to establish multimodal VA estimation as a soft biometric modality, showing across two identity-labeled video benchmarks that VA-trained backbone features implicitly encode identity-discriminative information complementary to the static appearance cues used by conventional face recognition, and that score-level fusion with ArcFace improves verification performance.
    
    \item We introduce Quality-Aware Gating (QAG), a learned per-sample reliability estimation mechanism that scales each modality's contribution during both training and inference based on estimated quality.
    
    \item We introduce Adaptive Modality Dropout (AMD), a quality-dependent regularization strategy that drops less reliable modalities more frequently during training, sharing quality estimates with QAG.
\end{enumerate}

The remainder reviews related work (Sec.~\ref{sec:related}), 
details the QAAF framework (Sec.~\ref{sec:method}), and presents empirical validation on Aff-wild2 followed by the biometric verification analysis on AFEW-VA and YTF that establishes VA features as a soft biometric modality (Sec.~\ref{sec:exp}).

\section{Related Work}
\label{sec:related}

\subsection{Multimodal Fusion for Emotion Recognition}

Multimodal fusion strategies for affective computing can be broadly categorized into early (feature-level), late (decision-level), and intermediate (also called hybrid) fusion~\cite{baltrusaitis2019multimodal}.
Transformer-based intermediate fusion approaches have become dominant, using cross-attention mechanisms to model inter-modal interactions~\cite{tsai2019multimodal, zhang2024joint}.
The Joint Multimodal Transformer (JMT)~\cite{zhang2024joint} aggregates information across modalities via multi-stream cross-attention, where each modality serves as the query for the other, followed by self-attention encoder layers and separate MLP regression heads for valence and arousal prediction.
While effective, JMT does not explicitly account for input quality variation, making it vulnerable to noisy or missing modalities.
We adopt cross-attention as our fusion primitive, following 
common practice in transformer-based multimodal methods, 
including a joint-representation stream as in JMT~\cite{zhang2024joint} for 
direct comparison. The core of QAAF, however, lies in two 
quality-aware mechanisms—Quality-Aware Gating and Adaptive 
Modality Dropout.

Recent methods on Aff-wild2~\cite{yu2025interactive, ahire2025maven, praveen2024recursive} achieve high Concordance Correlation Coefficient (CCC) through large end-to-end ensembles but do not explicitly address variable modality quality. Dresvyanskiy et al.~\cite{dresvyanskiy2024noisy} compared fusion strategies under noisy conditions but without quality-adaptive mechanisms. In contrast, the proposed QAAF framework directly models per-sample modality reliability and adapts the fusion mechanism accordingly.

\subsection{Handling Missing and Noisy Modalities}

Modality dropout~\cite{neverova2015moddrop} randomly zeros out modalities during training. Dai et al.~\cite{dai2024dropout} showed that uniform-probability dropout creates systematic bias, motivating quality-aware dropout. Recent works extend dropout with learnable tokens~\cite{contrastive2025moddrop} or prompts~\cite{guo2024prompt}, parameter-efficient adaptation~\cite{reza2025robust}, or missing-modality recovery via meta-learning~\cite{ma2021smil}, contrastive alignment~\cite{zhao2023missmodal}, or token reconstruction~\cite{nezakati2024mmp}.

Our approach differs fundamentally: rather than recovering missing information, we use a learned per-sample reliability score to gate each modality's contribution and modulate its dropout probability, requiring no explicit missing-modality labels or token reconstruction.

\subsection{Quality-Aware Multimodal Learning}

Quality-aware fusion has been explored primarily for classification tasks.
Zhang et al.~\cite{zhang2023qmf} proposed Quality-aware Multimodal Fusion (QMF), using energy-based uncertainty to estimate modality quality and adjust fusion weights, with theoretical generalization bounds.
Other quality-aware methods include Dirichlet-based belief estimation~\cite{cao2024pdf}, alternating unimodal adaptation~\cite{zhang2024mla}, entropy-based gating~\cite{agfn2025}, and gradient modulation~\cite{peng2022balanced}, but these target classification or operate only at inference time without quality-dependent training regularization.
Quality-conditioned aggregation has also been explored for unimodal face feature fusion under unconstrained conditions~\cite{jawade2023conan}, whereas QAAF extends this to a cross-modal audiovisual setting with explicit gating and dropout regularization. Conventional face image quality assessment~\cite{babnik2023diffiqa} operates at the input-image level for face recognition, whereas our quality estimator operates at the backbone-feature level driven by the VA training signal.

Our work differs from prior quality-aware methods in three key ways:
(1)~we combine quality estimation with both soft gating and adaptive dropout, providing complementary benefits during training and inference, whereas existing methods use quality for fusion weights or dropout, but not both;
(2)~we target continuous VA regression (optimized with CCC loss) rather than classification, where quality estimation interacts differently with the loss landscape;
(3)~we provide direct empirical comparison against QMF~\cite{zhang2023qmf}, demonstrating that learned quality estimation consistently outperforms energy-based estimation for regression tasks.

\subsection{Soft Biometrics and Affective Cues for Identity}

Soft biometric traits, such as facial actions, voice characteristics, and behavioral patterns, have long complemented conventional face recognition in unconstrained settings where static appearance alone is insufficient~\cite{dantcheva2016what,benedikt2010facial,kinnunen2010speaker}. Recent privacy-enhancement methods explicitly remove soft biometric attributes from face embeddings~\cite{rot2024aspecd}, implicitly assuming such attributes leak identity-discriminative information; our work demonstrates the converse: soft biometric structure emerges in features trained for affective rather than identity tasks. Prior soft-biometric approaches typically train explicitly for identity using discrete expression categories or hand-crafted dynamic descriptors. Our work differs in two respects: (i) we use continuous
valence--arousal as the training signal, capturing fine-grained expression dynamics; and (ii) identity-discriminative structure emerges implicitly as a byproduct of VA training, without any identity-specific supervision. To our knowledge, this is the first demonstration that audiovisual VA estimation yields representations usable as a soft biometric modality complementary to ArcFace-class face recognition. The persistence of identity-discriminative structure under emotion-centric fine-tuning is consistent with the view that VA training preserves rather than overwrites pretrained representations, mitigating catastrophic forgetting~\cite{nikhal2023mitigating} (Sec.~\ref{sec:biometric})---an effect that parallels the regularizing role of unsupervised and weakly supervised objectives in preserving general feature structure for biometric tasks~\cite{nikhal2023multicontext, nikhal2023weakly}.

Building on both the fusion challenges identified above and this soft biometric perspective, the next section presents the QAAF framework and the feature-based pipeline used for the identity verification analysis (1:1 matching).


\section{Method}
\label{sec:method}

\begin{figure*}[t]
\centering
\includegraphics[width=0.88\textwidth]{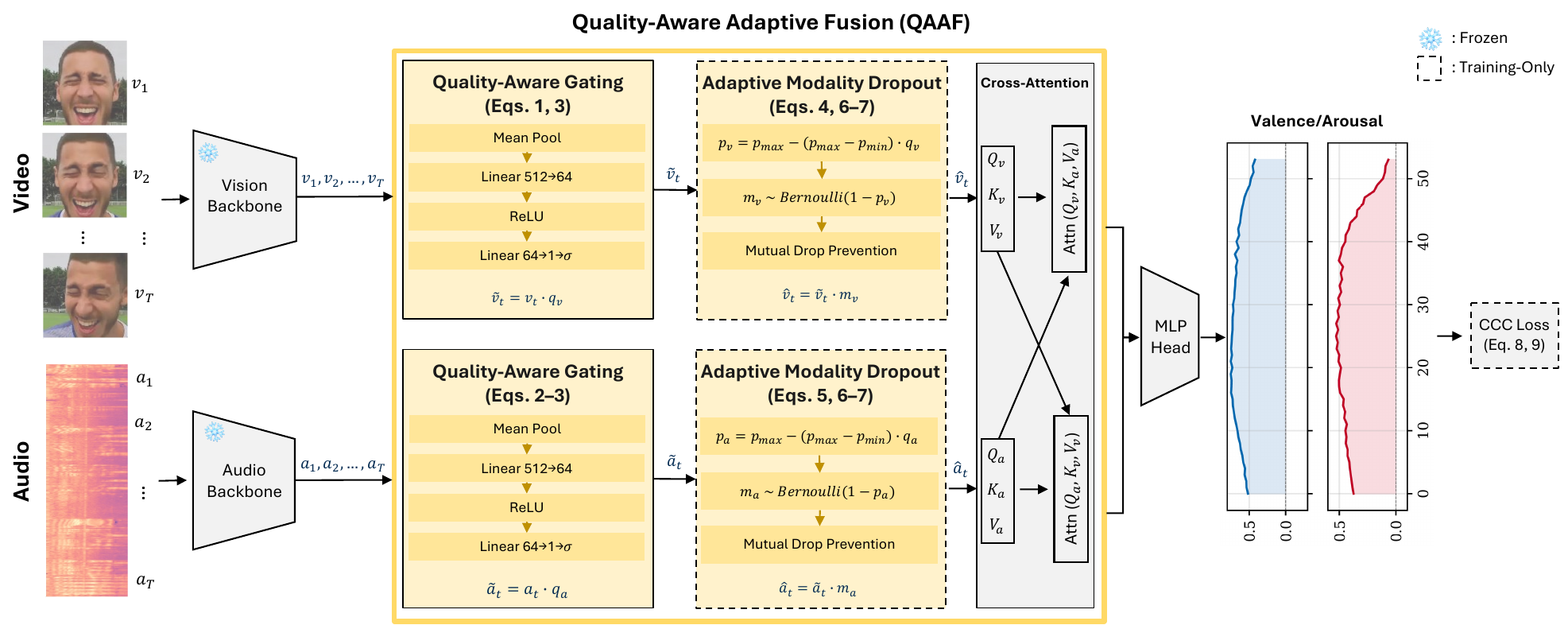}
\caption{Overview of the proposed Quality-Aware Adaptive Fusion (QAAF) framework. Quality estimators produce per-sample quality scores that drive both Quality-Aware Gating (QAG, active at train and inference) and Adaptive Modality Dropout (AMD, training only). The gated and dropped features are fused via cross-attention, followed by an MLP head for valence/arousal prediction. Snowflake denotes frozen modules; dashed borders denote training-only modules.}
\label{fig:framework}
\end{figure*}

\subsection{Overview}
Fig.~\ref{fig:framework} illustrates the proposed QAAF framework.
The complete pipeline processes raw video and audio inputs through the following stages: 
(1)~backbone networks extract visual and audio feature representations $\mathbf{V} \in \mathbb{R}^{B \times T \times D_v}$ and $\mathbf{A} \in \mathbb{R}^{B \times T \times D_a}$ from face frames and log-mel spectrograms (see Sec.~\ref{sec:backbones} for pre-processing details), where $B$ is the batch size, $T$ is the number of temporal steps, $D_v$, $D_a$ are the vision and audio feature dimensions, and $\mathbf{v}_t \in \mathbb{R}^{D_v}$, $\mathbf{a}_t \in \mathbb{R}^{D_a}$ denote the per-step features at $t = 1, \dots, T$;
(2)~independent learned linear projections map each modality to a common dimension $D{=}512$;
(3)~Quality-Aware Gating estimates per-sample modality reliability and applies soft gating;
(4)~Adaptive Modality Dropout provides reliability-based regularization (training only);
(5) cross-attention transformer fusion, where each modality serves as the query for the other ($Q$ from one modality, $K$ and $V$ from the other) along with a joint-representation stream, refined through self-attention encoder layers, produces fused representations;
(6)~separate regression heads predict valence and arousal.

\subsection{Backbone Networks}
\label{sec:backbones}

The framework uses backbone networks to extract feature representations from raw video frames and audio spectrograms.
The backbone weights are frozen after initial fine-tuning, and the QAAF fusion module is trained on the extracted features.

\textbf{Vision backbones.} Three architectures spanning CNNs and transformers are evaluated:
I3D~\cite{carreira2017quo} ($D_v{=}512$),
ViViT~\cite{arnab2021vivit} ($D_v{=}768$),
and VideoMAE~\cite{tong2022videomae} ($D_v{=}768$).
We use the Aff-wild2 official cropped-aligned face frames at $224\times224$, with each backbone's standard normalization (Kinetics-400 statistics for I3D; HuggingFace processors with their default statistics for ViViT and VideoMAE).
All vision backbones are fine-tuned on Aff-wild2, then frozen for feature extraction.

\textbf{Audio backbone.} ResNet18~\cite{he2016deep} fine-tuned on Aff-wild2 spectrograms is used as the audio backbone ($D_a{=}512$).
Audio is resampled to 44.1~kHz, and log-mel spectrograms are computed with 64 mel bins, a 20~ms Hann window, and a 10~ms hop length, then normalized via dataset-wide z-score statistics.

Features are projected to a common dimension $D{=}512$ via independent learned linear projections when backbone dimensions differ. This introduces an asymmetry across backbones---ViViT and VideoMAE undergo an additional projection from 768 to 512, while I3D and the audio backbone are used at their native dimensions. We discuss the implications in Sec.~\ref{sec:discussion}.

\subsection{Quality-Aware Gating (QAG)}
\label{sec:qag}

The core idea of QAG is to learn per-sample estimates of how much emotion-relevant information each modality carries, and scale contributions accordingly.
For each modality, a lightweight quality estimator $g_\theta$ takes the mean-pooled features as input and produces a scalar gate value:
\begin{align}
    q_v &= \sigma\!\left(g_{\theta_v}\!\left(\frac{1}{T}\sum_{t=1}^{T} \mathbf{v}_t\right)\right) \in [0, 1] \\
    q_a &= \sigma\!\left(g_{\theta_a}\!\left(\frac{1}{T}\sum_{t=1}^{T} \mathbf{a}_t\right)\right) \in [0, 1]
\end{align}
where $\sigma$ is the sigmoid function and $g_\theta: \mathbb{R}^D \to \mathbb{R}$ is a two-layer MLP ($D \to 64 \to 1$).

The gated features are computed via element-wise scaling:
\begin{equation}
    \tilde{\mathbf{v}}_t = q_v \cdot \mathbf{v}_t, \quad 
    \tilde{\mathbf{a}}_t = q_a \cdot \mathbf{a}_t 
    \quad \text{for } t = 1, \dots, T
\end{equation}

This soft gating is differentiable, allowing gradients from the task loss to flow directly to the quality estimators.
The model learns to assign higher gate values to informative modalities and lower values to noisy ones, without explicit quality supervision.

QAG is active during both training and inference, providing consistent quality-adaptive behavior.

\subsection{Adaptive Modality Dropout (AMD)}
\label{sec:amd}

Unlike the standard modality dropout~\cite{neverova2015moddrop}, which randomly zeros out entire modalities during training, the proposed Adaptive Modality Dropout uses quality-dependent probabilities that drop less reliable modalities more frequently:
\begin{align}
    p_v &= p_{\max} - (p_{\max} - p_{\min}) \cdot q_v \\
    p_a &= p_{\max} - (p_{\max} - p_{\min}) \cdot q_a
\end{align}
where $q_v, q_a \in [0, 1]$ are quality scores from QAG, $p_{\min}$ is the minimum dropout probability (applied to highest-quality inputs), and $p_{\max}$ is the maximum (applied to lowest-quality inputs). The bounds $p_{\min}$ and $p_{\max}$ are shared across both modalities.

For each sample in a batch, binary dropout masks are sampled:
\begin{align}
    m_v &\sim \text{Bernoulli}(1 - p_v), \quad
    m_a \sim \text{Bernoulli}(1 - p_a)
\end{align}

To prevent both modalities from being dropped simultaneously, if $m_v = m_a = 0$, the modality with higher estimated quality is retained:
\begin{align}
    m_v = \mathds{1}[q_v \geq q_a], \quad m_a = 1 - m_v
\end{align}
where $\mathds{1}[\cdot]$ denotes the indicator function, returning 1 when the condition is true and 0 otherwise. The dropped features used for cross-attention fusion are $\hat{\mathbf{v}}_t = m_v \cdot \tilde{\mathbf{v}}_t$ and $\hat{\mathbf{a}}_t = m_a \cdot \tilde{\mathbf{a}}_t$.

AMD operates only during training, serving as cross-modality regularization that teaches the model to recover the emotional target from partial observations. At inference, only QAG's soft gating is applied.

\subsection{Training Objective}

The primary training loss combines per-task CCC losses for valence and arousal:
\begin{align}
    \mathcal{L} = \mathcal{L}_{\text{CCC}}^{val} + \mathcal{L}_{\text{CCC}}^{aro},
\end{align}
where $\mathcal{L}^\tau_{CCC} = 1 - \text{CCC}(\hat{y}_\tau, y_\tau)$ for $\tau \in \{val, aro\}$.
The CCC is defined as:
\begin{align}
    \text{CCC}(\hat{y}, y) = \frac{2 \sigma_{\hat{y}y}}{\sigma_{\hat{y}}^2 + \sigma_y^2 + (\mu_{\hat{y}} - \mu_y)^2}
\end{align}
where $\sigma_{\hat{y}y}$ is the covariance between predictions and targets, $\sigma_{\hat{y}}^2$ and $\sigma_y^2$ are their respective variances, and $\mu_{\hat{y}}$ and $\mu_y$ are their means.

No additional loss terms are required for the quality estimators. They are trained jointly through the task loss via the differentiable gating mechanism.

\section{Experiments}
\label{sec:exp}

\subsection{Datasets}

\textbf{Aff-wild2}~\cite{kollias2019deep} is the largest in-the-wild dataset for continuous VA estimation, containing 594 videos from YouTube with frame-level annotations of valence and arousal in $[-1, 1]$.
The official train/validation split (356/76 video sequences) is used with backbone features extracted at 8 temporal steps per video.

\textbf{AFEW-VA}~\cite{kossaifi2017afew} and \textbf{YTF}~\cite{wolf2011face} are used for the biometric verification analysis (Sec.~\ref{sec:biometric}).
AFEW-VA contains 600 video clips from 67 actors with continuous VA annotations, providing a controlled setting to evaluate whether VA-trained features encode identity-discriminative information.
YTF (YouTube Faces DB) contains 3,425 videos of 1,595 subjects with the standard 5,000-pair verification protocol, offering a larger-scale in-the-wild identity benchmark to evaluate the generalization of this finding. \rev{Here, generalization refers to cross-dataset generalization: whether the identity-discriminative structure observed on AFEW-VA (67 actors) persists on YTF, a larger and independently collected population (1,595 subjects). We do not claim generalization to very large operational populations, which we leave to future work.}

\subsection{Implementation Details}

All \rev{fusion} experiments use unified hyperparameters: batch size 64, SGD with learning rate $10^{-4}$ and weight decay $10^{-4}$, base dropout 0.2, 4 attention heads, 2 layers per self-attention encoder block, and early stopping with patience 25. For AMD, $p_{min}{=}0.05$ and $p_{max}{=}0.4$. QAG adds only $\sim$66K parameters (0.4\% of 16.6M fusion parameters) with $<$2\% overhead. Each video is sampled to $T{=}8$ temporal steps. Results are reported as 10-seed averages (seeds 0--9), with statistical significance assessed via paired t-tests and Cohen's $d$ for effect size; we observed up to $\pm 0.1$ variation in Avg CCC (mean of CCC-Valence and CCC-Arousal) across runs.

\rev{Prior to fusion, each backbone is fine-tuned on Aff-wild2 with the CCC loss and then frozen for feature extraction. We use AdamW (learning rate $10^{-5}$) for ViViT and VideoMAE, SGD (learning rate $10^{-3}$) for I3D, and Adam (learning rate $10^{-4}$) for the audio backbone, selecting the checkpoint with the best validation CCC and applying no data augmentation. For the identity verification analysis, backbone features are mean-pooled over time and compared via cosine similarity, with each method's decision threshold set at its own Equal Error Rate operating point.}

\subsection{Experimental Configurations}

Six configurations isolate each component's contribution:
(1) Baseline, JMT fusion without gating or dropout;
(2) QMF~\cite{zhang2023qmf}, energy-based quality estimation from L2
feature norms without learnable parameters; (3) QMF+AMD, energy-based
gating combined with our Adaptive Modality Dropout; (4) Fixed Dropout,
standard modality dropout~\cite{neverova2015moddrop} ($p{=}0.20$);
(5) AMD only, Adaptive Modality Dropout without gating; and
(6) QAG+AMD, the full proposed framework.

\subsection{Main Results on Aff-wild2}

\begin{table}[t]
\centering
\caption{Comparison of fusion strategies on Aff-wild2 (10-seed averages). \textbf{Bold} = highest Avg CCC per backbone. Statistical significance 
vs. Baseline: $^*p<0.05, ^{**}p<0.01, ^{***}p<0.001$.}
\label{tab:main_results}
\smallskip
\resizebox{\columnwidth}{!}{
\begin{tabular}{l l cccc}
\toprule
Backbone & Method & CCC-V & CCC-A & Avg CCC & $\pm$std \\
\midrule
\multirow{6}{*}{ViViT}
& Baseline        & 0.232 & 0.343 & 0.288 & 0.035 \\
& QMF~\cite{zhang2023qmf} & 0.232 & 0.343 & 0.288 & 0.035 \\
& QMF + AMD       & 0.265 & 0.361 & \textbf{0.313}$^{***}$ & 0.036 \\
& Fixed Dropout   & 0.245 & 0.348 & 0.296 & 0.034 \\
& AMD only        & 0.264 & 0.357 & 0.310$^{**}$ & 0.037 \\
& QAG + AMD       & 0.269 & 0.346 & 0.308$^{***}$ & 0.029 \\
\midrule
\multirow{6}{*}{VideoMAE}
& Baseline        & 0.239 & 0.507 & 0.373 & 0.040 \\
& QMF~\cite{zhang2023qmf} & 0.239 & 0.507 & 0.373 & 0.041 \\
& QMF + AMD       & 0.272 & 0.533 & \textbf{0.402}$^*$ & 0.025 \\
& Fixed Dropout   & 0.247 & 0.517 & 0.382 & 0.034 \\
& AMD only        & 0.266 & 0.528 & 0.397$^*$ & 0.031 \\
& QAG + AMD       & 0.260 & 0.528 & 0.394 & 0.030 \\
\midrule
\multirow{6}{*}{I3D}
& Baseline        & 0.225 & 0.260 & 0.242 & 0.056 \\
& QMF~\cite{zhang2023qmf} & 0.219 & 0.263 & 0.241 & 0.056 \\
& QMF + AMD       & 0.214 & 0.298 & 0.256 & 0.057 \\
& Fixed Dropout   & 0.265 & 0.291 & 0.278 & 0.050 \\
& AMD only        & 0.268 & 0.308 & \textbf{0.288}$^{**}$ & 0.061 \\
& QAG + AMD       & 0.284 & 0.288 & 0.286 & 0.077 \\
\bottomrule
\end{tabular}}
\end{table}

\begin{table}[t]
\centering
\caption{Missing modality robustness (10-seed averages). ``Normal'' = both modalities; ``V-only'' / ``A-only'' = one modality zeroed. \%Drop V (resp. \%Drop A) = relative degradation from Normal to V-only (resp. A-only), computed from unrounded values. \textbf{Bold} = best per backbone per column.}
\label{tab:missing_modality}
\smallskip
\resizebox{\columnwidth}{!}{
\begin{tabular}{ll ccccc}
\toprule
Backbone & Method & Normal & V-only & A-only & \%Drop V & \%Drop A \\
\midrule
\multirow{6}{*}{ViViT}
& Baseline    & .288 & .233 & .010 & 19.2 & 96.5 \\
& QMF         & .288 & .233 & .010 & 19.2 & 96.5 \\
& QMF+AMD     & .313 & \textbf{.284} & .033 & 9.3 & \textbf{89.5} \\
& Fixed Drop. & .296 & .271 & .019 & 8.6 & 93.5 \\
& AMD         & .310 & .282 & .030 & 9.2 & 90.3 \\
& QAG+AMD     & .308 & \textbf{.284} & \textbf{.032} & \textbf{7.5} & 89.6 \\
\midrule
\multirow{6}{*}{VideoMAE}
& Baseline    & .373 & .257 & .021 & 31.1 & 94.4 \\
& QMF         & .373 & .257 & .021 & 31.1 & 94.4 \\
& QMF+AMD     & .402 & .307 & \textbf{.088} & 23.6 & \textbf{78.1} \\
& Fixed Drop. & .382 & .334 & .030 & 12.7 & 92.1 \\
& AMD         & .397 & .350 & .045 & 11.9 & 88.7 \\
& QAG+AMD     & .394 & \textbf{.356} & .052 & \textbf{9.7} & 86.8 \\
\midrule
\multirow{6}{*}{I3D}
& Baseline    & .242 & .069 & .021 & 71.3 & 91.3 \\
& QMF         & .241 & .068 & .021 & 71.7 & 91.3 \\
& QMF+AMD     & .256 & \textbf{.203} & .036 & \textbf{20.6} & 85.9 \\
& Fixed Drop. & .278 & .167 & .119 & 40.0 & 57.2 \\
& AMD         & .288 & .159 & .172 & 44.8 & 40.3 \\
& QAG+AMD     & .286 & .188 & \textbf{.175} & 34.4 & \textbf{38.8} \\
\bottomrule
\end{tabular}}
\end{table}

Table~\ref{tab:main_results} presents the main comparison across three vision backbones with 10-seed averages and paired $t$-test significance.

(1)~AMD achieves statistically significant improvements on all three backbones ($p{<}0.05$) with large effect sizes ($d{=}0.78$--$1.21$). QAG+AMD reaches significance on ViViT ($p{<}0.001$, $d{=}1.61$); on VideoMAE and I3D it shows comparable improvements with medium-to-large effect sizes ($d{=}0.54$--$0.71$) but does not reach conventional significance. The largest absolute gain is AMD on I3D: +0.046 over baseline.

(2)~Energy-based gating provides no benefit; learned gating does.
QMF~\cite{zhang2023qmf} is statistically indistinguishable from baseline on all backbones ($\Delta{\approx}0$, $p{>}0.27$), demonstrating that energy-based quality estimation (which uses L2 feature norms without learnable parameters) is ineffective for continuous VA regression where quality variations are subtle and task-dependent.
In contrast, AMD and QAG+AMD achieve gains of +0.020--0.046 over baseline.
Although QMF+AMD achieves the highest Avg CCC on ViViT and VideoMAE, this gain is driven entirely by AMD's regularization; on I3D, QMF+AMD (0.256) falls well below QAG+AMD (0.286), indicating that energy-based estimates mislead adaptive dropout for weaker backbones.
QAG+AMD's full-modality Avg CCC is marginally below AMD-only (0.308 vs.\ 0.310 on ViViT), but QAG's contribution lies in missing-modality robustness (Sec.~\ref{sec:missing}; V-only drop 7.5--34.4\% vs.\ 9.2--44.8\% for AMD-only) and ensemble effectiveness (Sec.~\ref{sec:ensemble}; +0.046 over the AMD-only ensemble).

(3)~AMD and QAG+AMD reduce run-to-run variance on ViViT and VideoMAE (0.029--0.037 vs.\ 0.035--0.040 for baseline), with comparable variance on I3D (0.061--0.077 vs.\ 0.056), supporting the quality-aware design of QAAF.

\subsection{Missing Modality Robustness}
\label{sec:missing}

A critical advantage of quality-aware fusion is graceful degradation under modality unavailability. We evaluate by zeroing out audio (V-only) or video (A-only) features. Table~\ref{tab:missing_modality} reports results.

Energy-based gating (QMF) provides no robustness benefit, showing identical degradation to baseline. Adding AMD (QMF+AMD) alleviates this (e.g., I3D drop 71.7\% $\to$ 20.6\%), but inconsistently across backbones.

QAG+AMD provides the most consistent robustness, reducing V-only degradation to 7.5\% on ViViT, 9.7\% on VideoMAE, and 34.4\% on I3D, a 1.9--3.2$\times$ improvement over baseline. The V-only gap is statistically significant on VideoMAE (QAG+AMD 0.356 vs.\ QMF+AMD 0.307, $p$=0.024, $d$=0.86). QAG+AMD also achieves the highest A-only CCC on ViViT and I3D, with an 8$\times$ improvement on I3D (0.175 vs.\ 0.021).

The \%Drop columns reveal an inherent asymmetry: the model relies more heavily on visual features than on audio (A-only drops 86.8--89.6\% on ViViT/VideoMAE and 38.8\% on I3D, versus 7.5--34.4\% for V-only), reflecting the relative information content of facial expressions versus vocal prosody for VA estimation~\cite{kollias2019deep, tzirakis2017end}.

\subsection{Late Fusion Ensemble}
\label{sec:ensemble}

We evaluate late fusion ensembles combining predictions from models with different vision backbones, with per-dimension blending weights grid-searched (step 0.1). Table~\ref{tab:ensemble} shows the QAG+AMD ensemble achieves 0.472 Avg CCC, outperforming the AMD ensemble (0.426) by +0.046\rev{, and the baseline ensemble under the same setting (0.415) by +0.057}. The gap stems from QAG nearly doubling I3D's CCC-V at the ensemble-selected best seed (0.156$\rightarrow$0.309), making it a productive ensemble contributor rather than a liability.

\begin{table}[t]
\centering
\caption{Late fusion ensemble results (best seed per backbone).
``Single Best'' = best individual backbone model.}
\label{tab:ensemble}
\setlength{\tabcolsep}{4pt}
\renewcommand{\arraystretch}{0.9}
\footnotesize
\begin{tabular}{l ccc}
\toprule
Method & CCC-V & CCC-A & Avg CCC \\
\midrule
\rev{Baseline Ensemble} & \rev{0.323} & \rev{0.508} & \rev{0.415} \\
\midrule
AMD Single Best        & 0.246 & 0.528 & 0.387 \\
AMD Ensemble           & 0.320 & 0.531 & 0.426 \\
\midrule
QAG+AMD Single Best    & 0.258 & 0.523 & 0.391 \\
QAG+AMD Ensemble       & \textbf{0.363} & \textbf{0.581} & \textbf{0.472} \\
\bottomrule
\end{tabular}
\end{table}

\subsection{Implicit Identity Information in VA Features}
\label{sec:biometric}

Audiovisual fusion has been directly applied to person verification with cross-attention mechanisms~\cite{praveen2024dca}. Building on this, we test whether VA estimation---without any identity supervision---yields identity-discriminative backbone representations via face verification on AFEW-VA (99 genuine, 6,534 impostor pairs) and YTF (5,000 pairs, standard protocol). \rev{For AFEW-VA, we build a gallery template for each actor by averaging that actor's training clips and match each test clip against it. For YTF, we match video pairs directly following the standard protocol.} Since backbones are frozen after Aff-wild2 fine-tuning and shared across fusion configurations, verification results depend only on VA fine-tuning, independent of QAG+AMD; we refer to them as VA-trained ViViT and VA-trained VideoMAE.

\begin{table}[t]
\centering
\caption{Verification benchmark on AFEW-VA (67 actors) and YTF 
(1,595 subjects). \textit{ViViT/VideoMAE pretrained} rows show Kinetics-400 pretrained backbones without VA fine-tuning, included as controls for the contribution of VA training. Rows grouped by type and sorted by AFEW-VA EER within each group. \textbf{Bold} = best per category per metric; \textit{italic} = our methods.}
\label{tab:verification}
\smallskip
\resizebox{\columnwidth}{!}{
\begin{tabular}{@{}ll cc cc@{}}
\toprule
& & \multicolumn{2}{c}{\textbf{AFEW-VA}} & \multicolumn{2}{c}{\textbf{YTF}} \\
\cmidrule(lr){3-4} \cmidrule(lr){5-6}
\textbf{Type} & \textbf{Method} & \textbf{EER} $\downarrow$ & \textbf{AUC} $\uparrow$ & \textbf{EER} $\downarrow$ & \textbf{AUC} $\uparrow$ \\
\midrule
\multirow{5}{*}{\textit{Hard}}
  & ArcFace~\cite{deng2019arcface}              & \textbf{0.022} & \textbf{0.990} & \textbf{0.106} & \textbf{0.957} \\
  & AdaFace~\cite{kim2022adaface}      & 0.097          & 0.971          & 0.242          & 0.824 \\
  & SFace~\cite{zhong2021sface}                 & 0.222          & 0.851          & 0.300          & 0.764 \\
  & EdgeFace~\cite{george2024edgeface}          & 0.306          & 0.763          & 0.228          & 0.841 \\
  & SynthDistill~\cite{boutros2023synthdistill} & 0.396          & 0.691          & 0.280          & 0.785 \\
\midrule
\multirow{3}{*}{\textit{Hard+Soft}}
  & ArcFace + VA-trained ViViT   & \textbf{0.021} & \textbf{0.990} & \textbf{0.104} & \textbf{0.958} \\
  & AdaFace + VA-trained ViViT  & 0.092          & 0.973          & 0.222          & 0.856 \\
  & EdgeFace + VA-trained ViViT & 0.166          & 0.880          & 0.214          & 0.865 \\
\midrule
\multirow{11}{*}{\textit{Soft}}
  & \textit{VA-trained ViViT (ours)}             & \textbf{0.166} & 0.880          & \textbf{0.291} & \textbf{0.787} \\
  & JMT R2D1~\cite{zhang2024joint}              & 0.182          & 0.869          & 0.315          & 0.751 \\
  & EmotiEffLib~\cite{savchenko2023emotiefflib} & 0.192          & \textbf{0.901} & 0.379          & 0.670 \\
  & JMT I3D~\cite{zhang2024joint}               & 0.264          & 0.813          & 0.480          & 0.528 \\
  & FER~\cite{savchenko2022fer}                 & 0.271          & 0.799          & 0.455          & 0.563 \\
  & VideoMAE pretrained~\cite{tong2022videomae} & 0.306          & 0.749          & 0.443          & 0.586 \\
  & \textit{VA-trained VideoMAE (ours)}          & 0.315          & 0.749          & 0.438          & 0.598 \\
  & ViViT pretrained~\cite{arnab2021vivit}      & 0.316          & 0.784          & 0.326          & 0.748 \\
  & DAN~\cite{wen2023dan}                       & 0.323          & 0.736          & 0.373          & 0.682 \\
  & POSTER++~\cite{mao2024posterv2}             & 0.355          & 0.680          & 0.445          & 0.579 \\
  & MAE-DFER~\cite{sun2023maedfer}              & 0.433          & 0.614          & 0.440          & 0.587 \\
\bottomrule
\end{tabular}}
\end{table}

Table~\ref{tab:verification} reports the results on both datasets alongside established face recognition and emotion recognition models. On AFEW-VA, the VA-trained ViViT backbone achieves an Equal Error Rate (EER) of 0.166 and Area Under the Curve (AUC) of 0.880, ranking first among evaluated emotion-recognition baselines (e.g., EmotiEffLib~\cite{savchenko2023emotiefflib}, 0.192 EER) and surpassing the lighter face recognition model EdgeFace~\cite{george2024edgeface} (0.306 EER), while remaining below state-of-the-art face recognition (ArcFace, AdaFace) as expected for a soft biometric signal. On YTF, VA-trained ViViT again ranks first in the soft category (EER 0.291, AUC 0.787). ArcFace~\cite{deng2019arcface} (0.022/0.106 EER on AFEW-VA/YTF) thus serves as a strong reference rather than a direct competitor.

To localize where the identity signal resides, we evaluate face verification at each stage of the VA model: backbone (768-dim), projected features (512-dim), encoder output (512-dim), and VA output (2-dim). The backbone row is shared across fusion configurations; other rows use the QAG+AMD single-best model.

Table~\ref{tab:layer_eer} shows that identity discrimination degrades progressively toward the VA prediction (e.g., 0.166$\to$0.465 on ViViT/AFEW-VA, 0.290$\to$0.449 on YTF), with final-stage EERs near the chance level of 0.500. VideoMAE shows smaller stage-to-stage changes, consistent with its limited response to VA fine-tuning. The degradation is driven by the $768{\to}2$ compression rather than temporal pooling: 768-dim features yield similar EER whether mean-pooled (0.166) or sequential (0.172), while the 2-dim VA output is near-random in both settings (0.454/0.459). VA trajectory signatures (statistics and histograms of the predicted sequence) also fail verification (best EER 0.421).

These results reveal a disentangled dual representation: the backbone embeds identity-discriminative structure as a byproduct of face-centered fine-tuning, while the task head filters it out to yield an identity-invariant emotion prediction. The VA output is therefore privacy-preserving by construction, while the backbone features provide a soft biometric signal orthogonal to static face recognition.

\begin{table}[t]
\centering
\caption{Layer-by-layer verification EER on AFEW-VA and YTF. Chance level: EER = 0.500.}
\label{tab:layer_eer}
\setlength{\tabcolsep}{5pt}
\renewcommand{\arraystretch}{0.95}
\small
\resizebox{\columnwidth}{!}{
\begin{tabular}{@{}ll c cc cc@{}}
\toprule
& & & \multicolumn{2}{c}{\textbf{AFEW-VA}} & \multicolumn{2}{c}{\textbf{YTF}} \\
\cmidrule(lr){4-5} \cmidrule(lr){6-7}
\textbf{Config} & \textbf{Stage} & \textbf{Dim} & \textbf{ViViT} & \textbf{VideoMAE} & \textbf{ViViT} & \textbf{VideoMAE} \\
\midrule
Shared  & Backbone           & 768 & 0.166 & 0.315 & 0.290 & 0.438 \\
\midrule
\multirow{3}{*}{QAG+AMD}
        & After projection   & 512 & 0.212 & 0.329 & 0.295 & 0.437 \\
        & After encoder      & 512 & 0.292 & 0.333 & 0.331 & 0.439 \\
        & VA output          & 2   & 0.465 & 0.465 & 0.449 & 0.476 \\
\bottomrule
\end{tabular}}
\end{table}

To isolate the contribution of VA training, we compare Kinetics-400 pretrained backbones against their VA-trained counterparts (Table~\ref{tab:verification}). ViViT improves substantially with VA fine-tuning ($0.316 \to 0.166$ EER on AFEW-VA, $0.326 \to 0.291$ on YTF), evidencing that VA training reorganizes features into an identity-discriminative representation. VideoMAE's Kinetics-400 features already encode identity at a comparable level, showing essentially no change after fine-tuning.

\begin{figure}[t]
\centering
\includegraphics[width=\columnwidth]{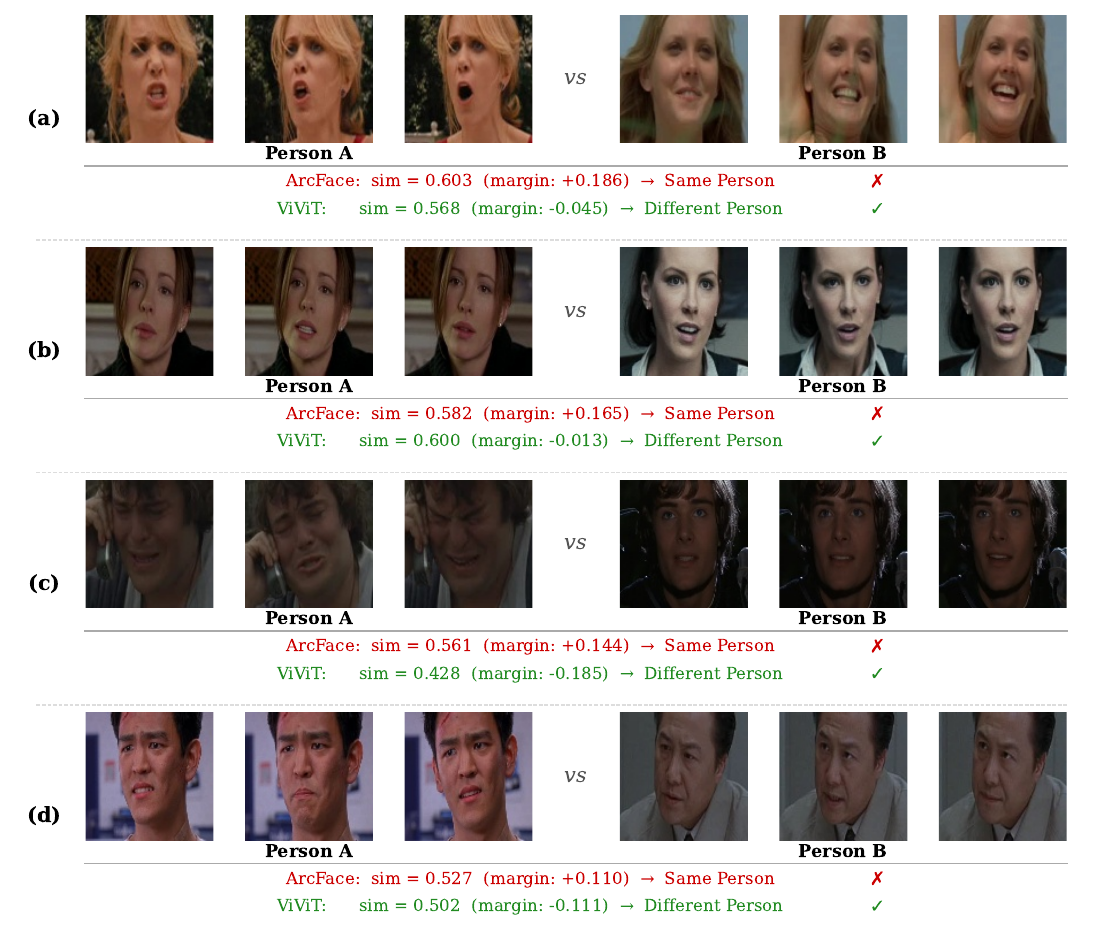}
\caption{Complementary verification examples. Each row shows a pair of different actors that ArcFace incorrectly matches (false accept) but the VA-trained ViViT correctly distinguishes. Similarity scores and margins relative to each model's threshold are shown.}
\label{fig:complementary}
\end{figure}

Fig.~\ref{fig:complementary} shows cases where ArcFace false-accepts distinct individuals but VA-trained ViViT correctly distinguishes them. Across all 148 ArcFace false accepts at its operating threshold, ViViT corrects 68.2\% (101 cases). Since ArcFace relies on static appearance, these corrections come from cues orthogonal to it, consistent with expression-dynamic identity structure.

To quantify this complementarity, we score-level fuse ArcFace and VA-trained ViViT cosine similarities (z-score normalized, $\alpha$ grid-searched over $[0,1]$). The best configuration reaches 0.021 EER on AFEW-VA ($\alpha{=}0.95$) and 0.104 EER on YTF ($\alpha{=}0.90$), improving over ArcFace alone. The optimal $\alpha$ is lower on YTF, where ArcFace operates further from saturation, suggesting VA features contribute more when the hard baseline has room for improvement. Fusion with AdaFace and EdgeFace (Table~\ref{tab:verification}) yields consistent complementary improvements on YTF.

\section{Discussion}
\label{sec:discussion}

Our central finding is that VA-trained representations encode identity-discriminative information complementary to static appearance, strong enough to improve a state-of-the-art face recognizer through simple score-level fusion, suggesting expression dynamics and prosody constitute a soft biometric modality worth investigating beyond affective computing.

QAAF also provides a lightweight VA estimation method: while end-to-end approaches~\cite{yu2025interactive, ahire2025maven, praveen2024recursive} achieve higher absolute CCC via large backbone ensembles with full fine-tuning, our feature-based ensemble (0.472 Avg CCC) operates with frozen backbones and $<$2\% overhead (Table~\ref{tab:sota}), offering a practical alternative when full fine-tuning is infeasible.

Several limitations qualify these results. The identity-encoding effect is backbone-dependent: VA fine-tuning substantially reduces EER for ViViT but not VideoMAE, whose Kinetics-400 features already encode identity. Validation on operational biometric deployments with more challenging conditions is left for future work. The absolute EER improvement over ArcFace is small due to its near-saturation; the 68.2\% correction rate on AFEW-VA false accepts is the more meaningful indicator of complementarity.

We also note an inherent asymmetry in the backbone comparison: ViViT and VideoMAE features ($D_v{=}768$) are projected to 512 dimensions to match I3D and audio, as the opposite design would instead penalize native-512 backbones. Still, QAAF's gains over baseline are consistent across backbones (+0.020/+0.021/+0.044 on ViViT/VideoMAE/I3D; Table~\ref{tab:main_results}) and so are missing-modality improvements (1.9--3.2$\times$; Table~\ref{tab:missing_modality}), so our conclusions are not driven by this asymmetry.

\begin{table}[t]
\centering
\caption{Comparison with recent Aff-wild2 VA estimation methods and computational overhead. $^\dagger$End-to-end methods with backbone fine-tuning. $^\ddagger$Feature-based (frozen backbones, fusion only).}
\label{tab:sota}
\smallskip
\resizebox{\columnwidth}{!}{
\begin{tabular}{ll ccc}
\toprule
\textbf{Type} & \textbf{Method} & \textbf{CCC-V} & \textbf{CCC-A} & \textbf{Avg CCC} \\
\midrule
\multirow{3}{*}{\shortstack[l]{\textit{End-to-end}$^\dagger$}}
  & Yu et al.~\cite{yu2025interactive}          & ---            & ---            & 0.600 \\
  & Ahire et al.~\cite{ahire2025maven}          & 0.596          & 0.683          & 0.640 \\
  & Praveen \& Alam~\cite{praveen2024recursive} & ---            & ---            & 0.545 \\
\midrule
\multirow{3}{*}{\shortstack[l]{\textit{Feature-based}$^\ddagger$}}
  & JMT Baseline~\cite{zhang2024joint}          & 0.232          & 0.343          & 0.288 \\
  & QAAF (ours, single best)                    & 0.269          & 0.346          & 0.308 \\
  & QAAF (ours, ensemble)                       & \textbf{0.363} & \textbf{0.581} & \textbf{0.472} \\
\bottomrule
\end{tabular}}
\end{table}

\section{Conclusion}
\label{sec:conclusion}

We have shown that audiovisual valence-arousal estimation, used as a pretext task, yields backbone representations that implicitly encode identity-discriminative information complementary to static appearance. On AFEW-VA and YTF, VA-trained ViViT features rank first among evaluated soft biometric methods without identity-specific training, and score-level fusion with ArcFace lowers EER on both datasets while correcting 68.2\% of ArcFace's false accepts on AFEW-VA. To learn these representations robustly, we proposed Quality-Aware Adaptive Fusion (QAAF), which estimates per-sample modality reliability via learned soft gating and quality-dependent dropout, \rev{reaching 0.472 Avg CCC on Aff-wild2 via late-fusion ensembling (0.415 for the baseline ensemble under the same setting)}. These findings establish multimodal VA estimation as a soft biometric modality complementary to conventional face recognition.


\section*{Acknowledgements}
This research was supported in part by the Maryland Governor's Office of Crime Prevention and Policy under Award No. PACT20260028. The views and conclusions expressed herein are those of the authors and should not be interpreted as necessarily representing the official policies, either expressed or implied, of the Maryland Governor's Office of Crime Prevention and Policy. The authors thank Prof. Shuvra Bhattacharyya and Prof. Kiminori Nakamura (University of Maryland, College Park) for valuable discussions and their ongoing collaboration on related research topics.

{\small
\bibliographystyle{ieee}
\bibliography{egbib}

@inproceedings{ringeval2013introducing,
  author    = {F. Ringeval and A. Sonderegger and J. Sauer and D. Lalanne},
  title     = {Introducing the {RECOLA} multimodal corpus of remote collaborative and affective interactions},
  booktitle = {Proc. IEEE FG},
  year      = {2013}
}

@article{kollias2019deep,
  author  = {D. Kollias and P. Tzirakis and M. A. Nicolaou and A. Papaioannou and G. Zhao and B. Schuller and I. Kotsia and S. Zafeiriou},
  title   = {Deep affect prediction in-the-wild: {Aff-wild} database and challenge, deep architectures, and beyond},
  journal = {IJCV},
  volume  = {127},
  pages   = {907--929},
  year    = {2019}
}

@inproceedings{zhang2024joint,
  author    = {P. Waligora and M. H. Aslam and M. O. Zeeshan and S. Belharbi and A. L. Koerich and M. Pedersoli and S. Bacon and E. Granger},
  title     = {Joint multimodal transformer for emotion recognition in the wild},
  booktitle = {Proc. IEEE/CVF CVPRw},
  pages     = {4625--4635},
  year      = {2024}
}

@article{tzirakis2017end,
  author  = {P. Tzirakis and G. Trigeorgis and M. A. Nicolaou and B. Schuller and S. Zafeiriou},
  title   = {End-to-end multimodal emotion recognition using deep neural networks},
  journal = {IEEE JSTSP},
  volume  = {11},
  number  = {8},
  pages   = {1301--1309},
  year    = {2017}
}

@inproceedings{tsai2019multimodal,
  author    = {Y.-H. H. Tsai and S. Bai and P. P. Liang and J. Z. Kolter and L.-P. Morency and R. Salakhutdinov},
  title     = {Multimodal transformer for unaligned multimodal language sequences},
  booktitle = {Proc. ACL},
  year      = {2019}
}

@article{baltrusaitis2019multimodal,
  author  = {T. Baltrusaitis and C. Ahuja and L.-P. Morency},
  title   = {Multimodal machine learning: A survey and taxonomy},
  journal = {IEEE TPAMI},
  volume  = {41},
  number  = {2},
  pages   = {423--443},
  year    = {2019}
}

@article{neverova2015moddrop,
  author  = {N. Neverova and C. Wolf and G. Taylor and F. Nebout},
  title   = {{ModDrop}: Adaptive multi-modal gesture recognition},
  journal = {IEEE TPAMI},
  volume  = {38},
  number  = {8},
  pages   = {1692--1706},
  year    = {2016}
}

@inproceedings{ma2021smil,
  author    = {M. Ma and J. Ren and L. Zhao and S. Tulyakov and C. Wu and X. Peng},
  title     = {{SMIL}: Multimodal learning with severely missing modality},
  booktitle = {Proc. AAAI},
  year      = {2021}
}

@inproceedings{zhang2023qmf,
  author    = {Q. Zhang and H. Wu and C. Zhang and Q. Hu and H. Fu and Y. Zhou and B. Peng},
  title     = {Provable dynamic fusion for low-quality multimodal data},
  booktitle = {Proc. ICML},
  year      = {2023}
}

@inproceedings{peng2022balanced,
  author    = {Y. Peng and Z. Huang and Z. Zhu and Q. Hu},
  title     = {Balanced multimodal learning via on-the-fly gradient modulation},
  booktitle = {Proc. CVPR},
  year      = {2022}
}

@inproceedings{carreira2017quo,
  author    = {J. Carreira and A. Zisserman},
  title     = {Quo vadis, action recognition? {A} new model and the kinetics dataset},
  booktitle = {Proc. CVPR},
  year      = {2017}
}

@inproceedings{arnab2021vivit,
  author    = {A. Arnab and M. Dehghani and G. Heigold and C. Sun and M. Lucic and C. Schmid},
  title     = {{ViViT}: A video vision transformer},
  booktitle = {Proc. ICCV},
  year      = {2021}
}

@inproceedings{tong2022videomae,
  author    = {Z. Tong and Y. Song and J. Wang and L. Wang},
  title     = {{VideoMAE}: Masked autoencoders are data-efficient learners for self-supervised video pre-training},
  booktitle = {Proc. NeurIPS},
  year      = {2022}
}

@inproceedings{he2016deep,
  author    = {K. He and X. Zhang and S. Ren and J. Sun},
  title     = {Deep residual learning for image recognition},
  booktitle = {Proc. CVPR},
  year      = {2016}
}

@inproceedings{yu2025interactive,
  author    = {J. Yu and Y. Wang and L. Wang and Y. Zheng and S. Xu},
  title     = {Interactive multimodal framework with temporal modeling for emotion recognition},
  booktitle = {Proc. CVPRw (8th ABAW)},
  year      = {2025}
}

@inproceedings{ahire2025maven,
  author    = {S. Ahire and others},
  title     = {{MAVEN}: Multi-modal attention for valence-arousal emotion network},
  booktitle = {Proc. CVPRw (8th ABAW)},
  year      = {2025}
}

@inproceedings{dresvyanskiy2024noisy,
  author    = {D. Dresvyanskiy and M. Markitantov and J. Yu and H. Kaya and A. Karpov},
  title     = {Multi-modal arousal and valence estimation under noisy conditions},
  booktitle = {Proc. CVPRw (7th ABAW)},
  year      = {2024}
}

@inproceedings{contrastive2025moddrop,
  author    = {Y. Gu and K. Saito and J. Ma},
  title     = {Learning contrastive multimodal fusion with improved modality dropout for disease detection and prediction},
  booktitle = {Proc. MICCAI},
  pages     = {280--290},
  year      = {2025}
}

@inproceedings{guo2024prompt,
  author    = {J. Guo and others},
  title     = {Multimodal prompt learning with missing modalities for sentiment analysis and emotion recognition},
  booktitle = {Proc. ACL},
  year      = {2024}
}

@article{zhao2023missmodal,
  author  = {J. Zhao and others},
  title   = {{MissModal}: Increasing robustness to missing modality in multimodal sentiment analysis},
  journal = {TACL},
  year    = {2023}
}

@inproceedings{nezakati2024mmp,
  author    = {N. Nezakati and M. K. Reza and A. Patil and M. Solh and M. S. Asif},
  title     = {{MMP}: Towards robust multi-modal learning with masked modality projection},
  booktitle = {Proc. IEEE BigData},
  year      = {2025}
}

@inproceedings{cao2024pdf,
  author    = {B. Cao and Y. Xia and Y. Ding and C. Zhang and Q. Hu},
  title     = {Predictive dynamic fusion},
  booktitle = {Proc. ICML},
  year      = {2024}
}

@article{agfn2025,
  author  = {Y. Chen and others},
  title   = {Beyond simple fusion: Adaptive gated fusion for robust multimodal sentiment analysis},
  journal = {arXiv:2510.01677},
  year    = {2025}
}

@inproceedings{dai2024dropout,
  author    = {Y. Dai and others},
  title     = {A study of dropout-induced modality bias on robustness to missing video frames for audio-visual speech recognition},
  booktitle = {Proc. CVPR},
  year      = {2024}
}

@article{reza2025robust,
  author  = {M. K. Reza and A. Prater-Bennette and M. S. Asif},
  title   = {Robust multimodal learning with missing modalities via parameter-efficient adaptation},
  journal = {IEEE TPAMI},
  volume  = {47},
  number  = {2},
  pages   = {1503--1518},
  year    = {2025}
}

@inproceedings{praveen2024recursive,
  author    = {R. G. Praveen and J. Alam},
  title     = {Recursive joint cross-modal attention for multimodal fusion in dimensional emotion recognition},
  booktitle = {Proc. CVPRw (6th ABAW)},
  year      = {2024}
}

@inproceedings{zhang2024mla,
  author    = {X. Zhang and J. Yoon and M. Bansal and H. Yao},
  title     = {Multimodal representation learning by alternating unimodal adaptation},
  booktitle = {Proc. CVPR},
  year      = {2024}
}

@inproceedings{deng2019arcface,
  author    = {J. Deng and J. Guo and N. Xue and S. Zafeiriou},
  title     = {{ArcFace}: Additive angular margin loss for deep face recognition},
  booktitle = {Proc. CVPR},
  year      = {2019}
}

@inproceedings{kim2022adaface,
  author    = {M. Kim and A. K. Jain and X. Liu},
  title     = {{AdaFace}: Quality adaptive margin for face recognition},
  booktitle = {Proc. CVPR},
  year      = {2022}
}

@article{george2024edgeface,
  author  = {A. George and C. Ecabert and H. O. Shahreza and K. Kotwal and S. Marcel},
  title   = {{EdgeFace}: Efficient face recognition model for edge devices},
  journal = {IEEE Trans. Biometrics, Behavior, and Identity Science},
  volume  = {6},
  number  = {2},
  pages   = {158--168},
  year    = {2024}
}

@inproceedings{savchenko2023emotiefflib,
  author    = {A. V. Savchenko},
  title     = {Facial expression recognition with adaptive frame rate based on multiple testing correction},
  booktitle = {Proc. ICML},
  year      = {2023}
}

@article{savchenko2022fer,
  author  = {A. V. Savchenko and L. V. Savchenko and I. Makarov},
  title   = {Classifying emotions and engagement in online learning based on a single facial expression recognition neural network},
  journal = {IEEE Trans. Affective Computing},
  volume  = {13},
  number  = {4},
  pages   = {2132--2143},
  year    = {2022}
}

@article{wen2023dan,
  author  = {Z. Wen and W. Lin and T. Wang and G. Xu},
  title   = {Distract your attention: Multi-head cross attention network for facial expression recognition},
  journal = {Biomimetics},
  volume  = {8},
  number  = {2},
  pages   = {199},
  year    = {2023}
}

@article{mao2024posterv2,
  author  = {Q. Mao and others},
  title   = {{POSTER++}: A simpler and stronger facial expression recognition network},
  journal = {Pattern Recognition},
  volume  = {149},
  pages   = {110292},
  year    = {2024}
}

@inproceedings{sun2023maedfer,
  author    = {L. Sun and others},
  title     = {{MAE-DFER}: Efficient masked autoencoder for self-supervised dynamic facial expression recognition},
  booktitle = {Proc. ACM MM},
  year      = {2023}
}

@article{kossaifi2017afew,
  author  = {J. Kossaifi and G. Tzimiropoulos and S. Todorovic and M. Pantic},
  title   = {{AFEW-VA} database for valence and arousal estimation in-the-wild},
  journal = {Image and Vision Computing},
  volume  = {65},
  pages   = {23--36},
  year    = {2017}
}

@article{zhong2021sface,
  author  = {Y. Zhong and W. Deng and J. Hu and D. Zhao and X. Li and D. Wen},
  title   = {{SFace}: Sigmoid-constrained hypersphere loss for robust face recognition},
  journal = {IEEE Trans. Image Processing},
  volume  = {30},
  pages   = {2587--2598},
  year    = {2021}
}

@inproceedings{boutros2023synthdistill,
  author    = {F. Boutros and M. Huber and P. Siebke and T. Riber and N. Damer},
  title     = {{SynthDistill}: Face recognition with knowledge distillation from synthetic data},
  booktitle = {Proc. IJCB},
  year      = {2023}
}

@article{benedikt2010facial,
  author  = {L. Benedikt and D. Cosker and P. L. Rosin and D. Marshall},
  title   = {Assessing the uniqueness and permanence of facial actions for use in biometric applications},
  journal = {IEEE Trans. Systems, Man, and Cybernetics --- Part A: Systems and Humans},
  volume  = {40},
  number  = {3},
  pages   = {449--460},
  year    = {2010}
}

@article{kinnunen2010speaker,
  author  = {T. Kinnunen and H. Li},
  title   = {An overview of text-independent speaker recognition: From features to supervectors},
  journal = {Speech Communication},
  volume  = {52},
  number  = {1},
  pages   = {12--40},
  year    = {2010}
}

@article{dantcheva2016what,
  author={Dantcheva, Antitza and Elia, Petros and Ross, Arun},
  title={What else does your biometric data reveal? A survey on soft biometrics},
  journal={IEEE Transactions on Information Forensics and Security},
  volume={11}, number={3}, pages={441--467}, year={2016}
}

@inproceedings{wolf2011face,
  title={Face recognition in unconstrained videos with matched background similarity},
  author={Wolf, Lior and Hassner, Tal and Maoz, Itay},
  booktitle={CVPR 2011},
  pages={529--534},
  year={2011},
  organization={IEEE}
}

@inproceedings{nikhal2023mitigating,
  title={Mitigating Catastrophic Interference using Unsupervised 
         Multi-Part Attention for RGB-IR Face Recognition},
  author={Nikhal, K. and Uzuegbunam, N. and Kennedy, B. and Riggan, B. S.},
  booktitle={IEEE/CVF Conference on Computer Vision and Pattern 
             Recognition Workshops (CVPRW)},
  year={2023}
}

@article{nikhal2023multicontext,
  title={Multi-Context Grouped Attention for Unsupervised Person 
         Re-Identification},
  author={Nikhal, K. and Riggan, B. S.},
  journal={IEEE Transactions on Biometrics, Behavior, and Identity 
           Science},
  volume={5},
  number={2},
  pages={170--182},
  year={2023}
}

@inproceedings{nikhal2023weakly,
  title={Weakly Supervised Face and Whole Body Recognition in 
         Turbulent Environments},
  author={Nikhal, K. and Riggan, B. S.},
  booktitle={IEEE International Joint Conference on Biometrics (IJCB)},
  year={2023}
}

@inproceedings{jawade2023conan,
  author    = {Bhavin Jawade and Deen Dayal Mohan and Dennis Fedorishin
               and Srirangaraj Setlur and Venugopal Govindaraju},
  title     = {{CoNAN}: Conditional Neural Aggregation Network for Unconstrained
               Face Feature Fusion},
  booktitle = {Proc. IEEE International Joint Conference on Biometrics (IJCB)},
  pages     = {1--10},
  year      = {2023},
  doi       = {10.1109/IJCB57857.2023.10448797}
}

@inproceedings{babnik2023diffiqa,
  author    = {{\v{Z}}iga Babnik and Peter Peer and Vitomir {\v{S}}truc},
  title     = {{DifFIQA}: Face Image Quality Assessment Using Denoising
               Diffusion Probabilistic Models},
  booktitle = {Proc. IEEE International Joint Conference on Biometrics (IJCB)},
  pages     = {1--10},
  year      = {2023},
  doi       = {10.1109/IJCB57857.2023.10449044}
}

@inproceedings{praveen2024dca,
  author    = {R. Gnana Praveen and Jahangir Alam},
  title     = {Dynamic Cross Attention for Audio-Visual Person Verification},
  booktitle = {Proc. IEEE International Conference on Automatic Face and
               Gesture Recognition (FG)},
  pages     = {1--8},
  year      = {2024}
}

@inproceedings{rot2024aspecd,
  author    = {Peter Rot and Philipp Terh{\"o}rst and Peter Peer and
               Vitomir {\v{S}}truc},
  title     = {{ASPECD}: Adaptable Soft-Biometric Privacy-Enhancement Using
               Centroid Decoding for Face Verification},
  booktitle = {Proc. IEEE International Conference on Automatic Face and
               Gesture Recognition (FG)},
  pages     = {1--10},
  year      = {2024}
}
}

\end{document}